\documentclass[journal]{IEEEtran}
\usepackage{cite}
\ifCLASSINFOpdf

\else

\fi

\usepackage{amsfonts,amssymb}
\usepackage{graphicx}
\usepackage{amsmath}
\usepackage{booktabs}
\usepackage{multirow}
\usepackage{caption}
\usepackage{color}
\usepackage[switch]{lineno}
\usepackage{ragged2e}
\usepackage{paralist}
\usepackage{enumitem}

\usepackage{times}
\usepackage{amsmath}
\usepackage{amssymb}

\usepackage{subfigure}
\usepackage{cite}

\usepackage{url}
\usepackage{amsfonts,amssymb}
\usepackage{graphicx}
\usepackage{amsmath}
\usepackage{booktabs}
\usepackage{multirow}
\usepackage{caption}
\usepackage{color}
\usepackage[switch]{lineno}
\usepackage{ragged2e}
\usepackage{paralist}
\usepackage{enumitem}
\usepackage{lineno}

\usepackage{caption}
\usepackage{bm}
\usepackage{enumitem}
\usepackage[ruled,linesnumbered]{algorithm2e}
\usepackage{algorithmicx}

\newcommand{\ie}{\emph{i.e., }}
\newcommand{\eg}{\emph{e.g., }}

\newcommand{\wrt}{\emph{w.r.t. }}
\newcommand{\cf}{\emph{cf. }}

\usepackage{subfigure}
\usepackage{color}

\begin{document}

\title{Causal Incremental Graph Convolution for Recommender System Retraining}

\author{Sihao Ding, Fuli Feng, Xiangnan He, Yong Liao, Jun Shi, and Yongdong Zhang
\IEEEcompsocitemizethanks {\IEEEcompsocthanksitem
S. Ding, X He, Y Liao, Y Zhang are  with University of Science and Technology of China, Hefei 230026, China (email: Dsihao@mail.ustc.edu.cn, hexn, ly, zhyd73 @ustc.edu.cn). F Feng is with the National University of Singapore (email: fulifeng93@gmail.com). J Shi is with Innovation Lab of CETC (email: junshi@cyberaray.com). Corresponding author: Fuli Feng.}
\thanks{This work is supported by the National Natural Science Foundation of China (U19A2079, U21B2026, 62121002).}
}

\markboth{IEEE TRANSACTIONS ON Neural Networks and Learning Systems}
{S. Ding \MakeLowercase{\textit{et al.}}: Incremental Graph Convolution and Causal effect distilling for Recommender System Retraining}

\IEEEcompsoctitleabstractindextext{
\begin{abstract}
Real-world recommender system needs to be regularly retrained to keep with the new data. In this work, we consider how to efficiently retrain graph convolution network (GCN) based recommender models, which are state-of-the-art techniques for collaborative recommendation. To pursue high efficiency, we set the target as using only new data for model updating, meanwhile not sacrificing the recommendation accuracy compared with full model retraining. This is non-trivial to achieve, since the interaction data participates in both the graph structure for model construction and the loss function for model learning, whereas the old graph structure is not allowed to use in model updating. 

Towards the goal, we propose a \textit{Causal Incremental Graph Convolution} approach, which consists of two new operators named \textit{Incremental Graph Convolution} (IGC) and \textit{Colliding Effect Distillation} (CED) to estimate the output of full graph convolution. In particular, we devise simple and effective modules for IGC to ingeniously combine the old representations and the incremental graph and effectively fuse the long-term and short-term preference signals. CED aims to avoid the out-of-date issue of inactive nodes that are not in the incremental graph, which connects the new data with inactive nodes through causal inference. In particular, CED estimates the causal effect of new data on the representation of inactive nodes through the control of their collider. 
Extensive experiments on three real-world datasets demonstrate both accuracy gains and significant speed-ups over the existing retraining mechanism.
\end{abstract}

}

\maketitle

\IEEEdisplaynotcompsoctitleabstractindextext

\IEEEpeerreviewmaketitle

\section{INTRODUCTION}
\label{sec:intro}

Recent years have witnessed the success of GCN-based recommender models such as PinSAGE~\cite{pinsage} and LightGCN~\cite{LightGCN}, which perform node representation learning over the interaction graph and demonstrate promising performance~\cite{NGCF}. The core of them is \textit{neighborhood aggregation} which enhances a node's representation with the information from its neighbors. In this way, the graph structure can be explicitly integrated into the embedding space, improving the representations of users and items. In practical usage, a recommender system needs to be periodically (\eg daily) retrained to keep the model fresh with the new interaction data. In this work, we study the problem of GCN model retraining for recommendation, which has received relatively little scrutiny.

\begin{figure}[t]
	\centering
	\includegraphics[width=0.45\textwidth]{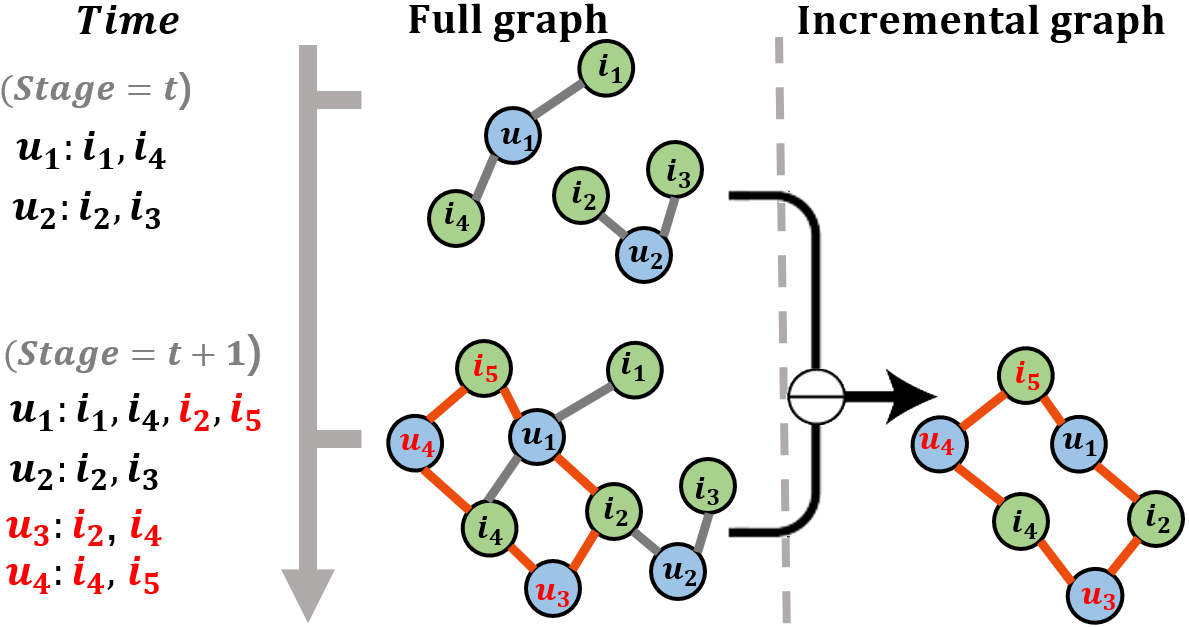}
 	\vspace{-7pt}
	\caption{An illustration of the incremental graph at stage $t+1$ in GCN retraining.}
	\label{fig:introduction graph}
 	\vspace{-11pt}
\end{figure}

Given new interactions for refreshing an old GCN model, there are three straightforward strategies:
\begin{itemize}[leftmargin=*]
    \item \textit{Full retraining}, which simply merges the old data and new interactions to perform a full model training. This solution retains the most fidelity since all data is used. However, it is very costly in both memory and time, since interaction data is usually of a large scale and keeps increasing with time.
    \item \textit{Fine-tuning with old graph.} Interaction data participates in two parts of a GCN: forming the graph structure to perform graph convolutions and constituting training examples of the loss function. This fine-tuning solution constructs training examples with new interactions only, while still uses the full graph structure. As such, although this solution costs fewer resources than the full retraining, it is still costly due to the usage of the old graph. 
    \item \textit{Fine-tuning w/o old graph.} This solution uses only the new interactions for model training and graph convolution. The old graph is not used in graph convolution, which saves many computation and storage resources because of the high sparsity of incremental graph (see Fig.~\ref{fig:introduction graph}). However, the new interactions contain only users' short-term preferences, which can differ much from the long-term performances and be much sparser. In addition, it cuts off the connection to the inactive nodes that have no new interactions. It thus suffers easily from forgetting and over-fitting issues. 
\end{itemize}

Given the pros and cons of the above intuitive strategies, we distill three considerations for effective and efficient GCN recommender retraining: 1) detaching the old graph; 2) reserving the old (long-term) preference signal; and 3) fusing the old and new preference signals. In short, our target is to achieve comparable or even better recommendation accuracy as full retraining, with the use of new interactions only. The key lies in how to estimate the output of \textit{full graph convolution} (\ie on the whole graph for full retraining), based on the \textit{old node representations} and the \textit{incremental graph} (\ie the graph constructed from the new interactions, \cf Fig. ~\ref{fig:introduction graph} as an example). This is however non-trivial to achieve for three reasons: 1) in the full graph convolution, the new interactions not only bring new neighbors for a target node, but also participates in the normalization weighting of old neighbors. It is known that the normalization weights of neighbors have a large impact on the GCN performance~\cite{LightGCN} and need to be carefully considered. 2) The old representations are learned over historical data and represent long-term preference. Since user interests may drift, making the new interactions discrepant from long-term preference, blindly fine-tuning old representations could make the model forget the long-term preference. 3) The incremental graph lacks inactive nodes that have no new interactions (\eg node $i_{1}$ in Fig.~\ref{fig:introduction graph}), which calls for extra effort to refresh the representation of such nodes.

Towards our target, we first propose \textit{Incremental Graph Convolution}, which estimates the full graph convolution of a target node based on its old representation and the incremental graph structure. In particular, we design a \textit{degree synchronizer} to learn the degree-based node normalization weights, so as to approach the normalization weights in full graph convolution. Meanwhile, we devise a \textit{representation aggregator} to compose the old representation with the new neighbors for effectively encoding both long-term and short-term signals. We devise the two modules as simple convolutional neural networks (CNNs), which keep the overall complexity of IGC close to fine-tuning w/o old graph with sufficient fidelity. 

Meanwhile, we devise \textit{Colliding Effect Distillation} to refresh the representation of inactive nodes, which estimates the effect of new data on the representation of inactive nodes. In particular, we resort to causal theory~\cite{why-book} and frame the whole incremental training phase as a causal graph (see Fig.~\ref{fig:causal graph}). The key lies in how to connect the $d$-separated~\cite{why-book} variables: the representation of inactive nodes and new data. We construct a collider~\cite{why-book} between them to make them conditionally $d$-connected and estimate their colliding effect~\cite{Distilling} to update the representation of inactive nodes.
In particular, CED distills the colliding effect conditioned on the similarity between inactive and active nodes implied by the old representations.

The proposed IGC and CED are universal operators that are applicable to most GCN models. In this work, we equip them on LightGCN~\cite{LightGCN}, a simple GCN model with state-of-the-art performance on collaborative recommendation. Through extensive experiments on three real-world datasets, we demonstrate the effectiveness of the Causal Incremental Graph Convolution approach, which outperforms full retraining in recommendation accuracy with overall costs comparable to fine-tuning w/o old graph. The main contributions of this work are summarized as follows:
\begin{itemize}[leftmargin=*]
    \item We study the new task of GCN model retraining for recommendation, and propose a new \textit{Causal Incremental Graph Convolution} approach that well supports the efficient and effective retraining of GCN models.
    \item We devise two universal operators named \textit{Incremental Graph Convolution} and \textit{Colliding Effect Distillation}, that can estimate the full graph convolution for both active and inactive nodes during the incremental training.
    \item We instantiate IGC and CED on LightGCN and conduct extensive experiments to demonstrate the effectiveness and efficiency of our approach. 
\end{itemize}

\section{METHODOLOGY}
\label{Problem formulation}

We represent the temporal interaction data as $\mathcal{I} = \left \{ I_{0}, I_{1}, \ldots, I_{t-1}, I_{t} \right \}$, and the corresponding user-item graph as $\mathcal{G} = \left \{ G_{0}, G_{1}, \ldots, G_{t-1}, G_{t} \right \}$. $I_{t}$ means the new interactions collected in stage $t$; $G_{t}$ is the bipartite user-item graph built on the interactions in $I_{t}$, which we also term as \textit{incremental graph}. A stage can be of any time period (\eg one/multiple hours/days), which depends on the expected model freshness and the training cost we can afford.
Towards the target of efficient GCN model retraining, we formulate the task as:
\begin{align}\small
\label{equ:incremental task}
	\bm{\theta}_{t-1} \xrightarrow[retrain]{\left(I_{t}, G_{t}\right)} \bm{\theta}_{t} \xrightarrow[serve]{} I_{t+1}, 
\end{align}
where $\bm{\theta}_t$ denotes the model parameters learned in stage $t$.

In this formulation, we retrain using only the latest stage's interactions $I_{t}$ and the incremental graph $G_{t}$\footnote{Real-world recommender systems will continually collect new data, so the whole training data and the full graph will keep growing. Thus, the existing gap in retraining costs between \textit{fine-tuning with old graph} (or \textit{full-retraining}) and \textit{incremental training} will continue to widen.}. The model with $\bm{\theta}_t$ is used to serve for next stage $t+1$, thus we use the (future) data $I_{t+1}$ to evaluate the retraining effectiveness. 

\subsection{GCN-based Recommender Model}
We recap how GCN works for collaborative filtering. Suppose we construct GCN on the \textit{full graph} up to $t$, denoted as $G_{0\sim t}$, the graph convolution commonly used is:
\begin{align}
\label{equ:lgcn}
    \bm{e}_{i, t}^{(l+1)} = \sigma \bigg( \frac{1}{\sqrt{d_{i, 0\sim t}}} \overbrace{\sum_{j\in \mathcal{N}_{i, 0\sim t}} \frac{1}{\sqrt{d_{j, 0\sim t}}} \bm{e}_{j, t}^{(l)} }^{\text{neighborhood aggregation}} \bigg),
\end{align}
where $\bm{e}_{j, t}^{(l)}$ denotes the representation of node $j$ at $l$-th layer, $\mathcal{N}_{i, 0\sim t}$ represents the neighbors of node $i$ in the graph $G_{0\sim t}$, and $d_{i, 0\sim t}$ (accumulated degree) equals to the number of nodes in $\mathcal{N}_{i, 0\sim t}$. The core of graph convolution is \textit{neighborhood aggregation}, which aggregates the representations of neighbor nodes for the target node $i$. The node degrees play the role of normalization, exerting a large impact on the GCN performance~\cite{LightGCN}. The feature transformation function $\sigma(\cdot)$ has various formats such as linear~\cite{NGCF} and bilinear mapping~\cite{BGCN-Zhu_HongMin}. We focus on the neighborhood aggregation in this work, omitting the feature transformation function $\sigma(\cdot)$ for briefness.

The embeddings of the $0$-th layer are the model parameters to learn, which are trained by minimizing the loss function:
\begin{align}\label{eq:loss}
    \sum_{(u, i) \in \mathcal{I} / \mathcal{I}^-} L\Big(y_{u,i}, \hat{y}_{u, i}\Big) + \lambda \|\bm{\theta}\|^2,
\end{align}
where $I^-$ are negative samples, $y_{u, i}$ is the interaction label, and $\hat{y}_{u, i}$ is the corresponding model prediction such as the inner product of final user and item representations. $L(\cdot)$ specifies the recommendation loss such as the pairwise BPR~\cite{BPRloss} and pointwise cross entropy~\cite{NCF}, and $\lambda$ is the hyper-parameter for $L_2$ regularization. 

\subsection{Incremental Graph Convolution}
\label{subsec:IGC-LightGCN}
\begin{figure}[t]
	\centering
	\includegraphics[width=0.42\textwidth]{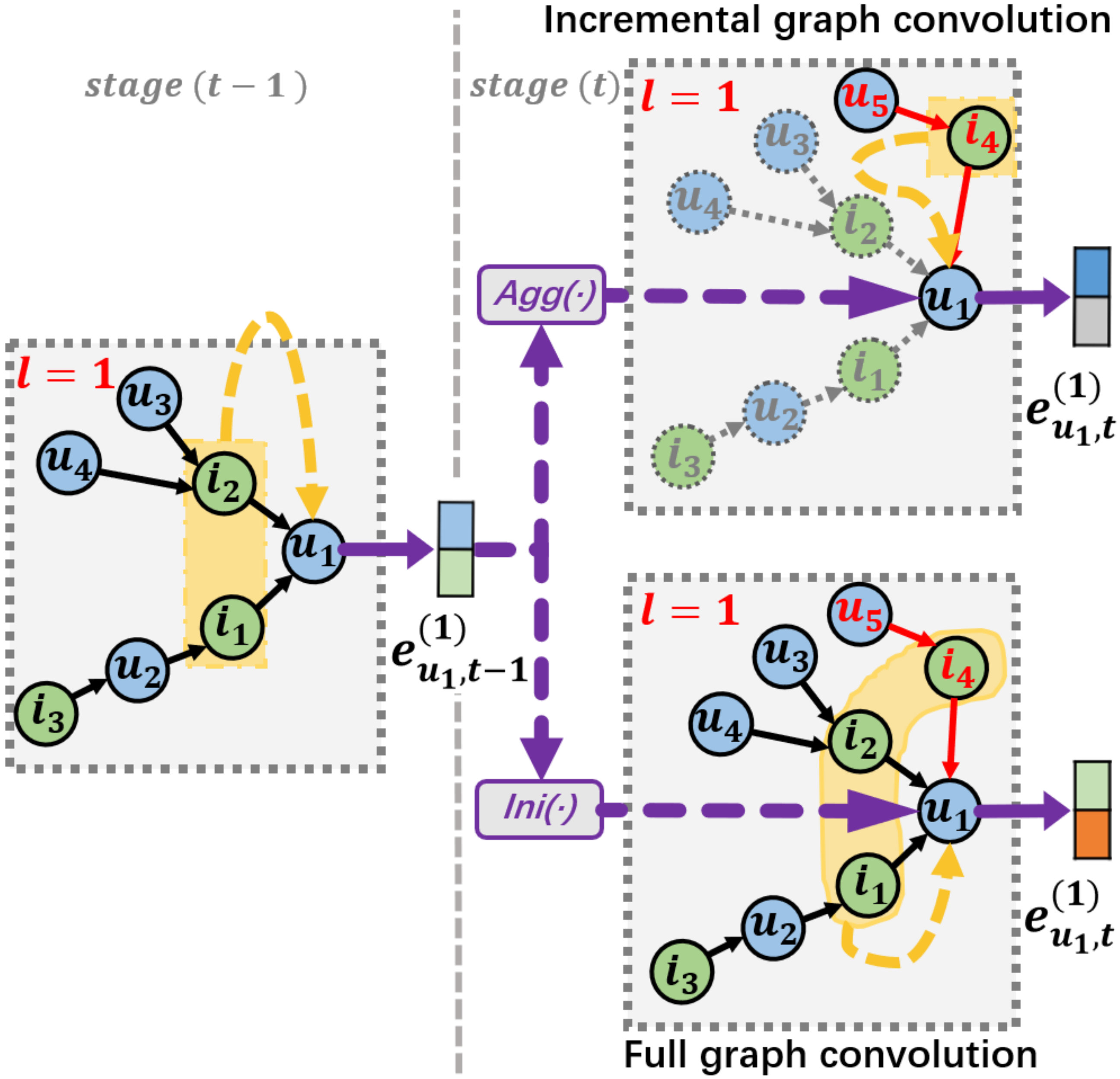}
	\vspace{-4pt}
	\caption{Incremental Graph Convolution on target node $u_1$ in stage $t$. The blurred part is the old graph that is not used in stage $t$. Red edges denote the new interactions in stage $t$. The yellow region shows the nodes included in graph convolution. $Agg(\cdot)$, $Ini(\cdot)$ represent the operations to use $\bm{e}_{u_1,t-1}^{(1)}$ in \textit{IGC} and \textit{full graph convolution}, respectively. }
	\vspace{-3pt}
    \label{fig:IGC-LightGCN}
\end{figure}
The cost of full retraining is high and increasing with time. Since $\mathcal{N}_{i,0\sim t}$ equals to $ \mathcal{N}_{i,0\sim t-1} \cup \mathcal{N}_{i,t}$, an efficient retraining means that bypassing $\mathcal{N}_{i,0\sim t-1}$ and using only $\mathcal{N}_{i,t}$ to estimate the full graph convolution of Eq.~(\ref{equ:lgcn}). Suppose the old representation $\bm{e}_{i,t-1}^{(l + 1)}$ is trained on $G_{0\sim t-1}$, it then can encode graph convolution experience on $\mathcal{N}_{i,0\sim t-1}$\footnote{Iteratively training with our IGC from stage $0$ to $t-1$ yields the same effect, so the premise still holds.}. As such, a smart integration of it with only new neighbors can well approximate the result of full graph convolution. To this end, we propose \textit{Incremental Graph Convolution}, which estimates the full graph convolution on target node $i$ as:
\begin{small}
\begin{align}
\label{lgcn pro}\small
    \bm{e}_{i,t}^{(l+1)} = \frac{1}{\sqrt{d_{i,t}'}} \cdot
        \varphi \bigg(
            \overbrace{\sqrt{d_{i, 0\sim t-1}} \cdot \bm{e}_{i,t-1}^{(l+1)}}^{\text{old representation, \textbf{constant}}}
            , \overbrace{\sum_{j\in \mathcal{N}_{i, t}} \frac{1}{\sqrt{d_{j, t}'}} \cdot \bm{e}_{j, t}^{(l)}}^{\text{new representations, \textbf{to learn}}}
        \bigg),
\end{align}
\end{small}
where $d_{j,t}' = f(d_{j, 0\sim t-1}, d_{j, t})$ denotes the normalization weight estimated by the \textit{degree synchronizer}. In the next, we elaborate the core designs of our IGC. 
\subsubsection{Degree Synchronizer}
To retain the most fidelity as the full graph convolution, we carefully set the degree-based normalization weights in IGC. We first scrutinize the first term of old representation: 
\begin{equation}
    \sqrt{d_{i, 0\sim t-1}} \cdot \bm{e}_{i,t-1}^{(l+1)} = \sum_{j\in \mathcal{N}_{i, 0\sim t-1}} \frac{1}{\sqrt{d_{j, 0\sim t-1}}} \bm{e}_{j, t-1}^{(l)}.
\end{equation}
If we replace $t-1$ with $t$, the right side becomes exactly what we need for full neighborhood aggregation. Since we are not allowed to use $\mathcal{N}_{i, 0\sim t-1}$ in IGC, we use $\sqrt{d_{i, 0\sim t-1}} \cdot \bm{e}_{i,t-1}^{(l+1)}$ instead, which has encoded the signal in $\mathcal{N}_{i, 0\sim t-1}$. As updating $\bm{e}_{i,t-1}^{(l+1)}$ in retraining has the risk of losing the signal of old neighbors, we set it as a constant during stage $t$ retraining. For this operation, we only need to store one additional integer for each node --- its accumulated degree $d_{i, 0\sim t-1}$, the cost of which is negligible. 

We next move to the second term for new neighbors modeling. At the first sight, we can set $f(d_{j, 0\sim t-1}, d_{j, t})$ as $d_{j, 0\sim t-1} + d_{j, t}$, which is the one used in full graph convolution. However, the evolution pattern of node degree may be different for datasets of different domains. This provides opportunities to boost the performance of original graph convolution, if we can strategically learn $f(\cdot)$ towards the recommendation objective function. To this end, we weighted combine the two degrees:
\begin{equation}
    f(d_{j, 0\sim t-1}, d_{j, t}) = \beta \cdot d_{j, 0\sim t-1} + d_{j, t},
\end{equation}
where $\beta$ is end-to-end trained to control the impact of old degree. We empirically find this simple function works well, and the optimal $\beta$ varies significantly for different datasets (\cf Section ~\ref{sec:ab study}). 

\subsubsection{Representation Aggregator $\varphi(\cdot)$}
The full graph convolution updates parameters associated with both old and new neighbors. However, IGC is only allowed to update parameters of new neighbors for efficiency concern. This causes significant discrepancy issues: 1) the old representation is off-the-shelf while the new representation is iteratively updated to optimize the loss function on new interactions, and 2) the old representation is learned over historical data and represents long-term preference, which may be discrepant from the new interactions due to interest drift.

To tackle these issues, we devise a parameterized function $\varphi(\cdot)$, adapting the fusion strategy to the training objective. An existing solution is the Transfer component in~\cite{SML}, which adopts a standard CNN network with nonlinear transformations. However, we find it works poorly in our scenario, partially because of the complexity brought by feature mapping and nonlinear activation. As such, we further simplify the CNN design, retaining the minimum operations and parameters. Specifically, we stack the old and new representations, passing them into a convolution layer:
\begin{align}\small
\label{IGC final}
    p
    \left(
            \bigg<
                \bm{w}_f^{(l)}, 
                \bigg[
                    \sqrt{d_{i, 0\sim t-1}} \cdot \bm{e}_{i,t-1}^{(l+1)}, \sum_{j\in \mathcal{N}_{i, t}} \frac{1}{\sqrt{d_{j, t}'}} \cdot \bm{e}_{j, t}^{(l)}
                \bigg]
            \bigg>
    \right),
\end{align}
where $p(\cdot)$ is a pooling operation, $\bm{w}_f \in \mathbb{R}^{2 \times 1}$ denotes the $f$-th filter of the CNN layer, and there can be multiple filters in a layer. In this way, the aggregator adjusts the importance of long-term and short-term preference and aligns the scale of the two representations in the training process. We can stack multiple CNN layers to enhance the expressiveness of the aggregator, whereas in our experiments using one layer leads to good performance in most cases (\cf Fig.~\ref{fig:RA-complexity}). \\

\noindent To summarize, compared with the vanilla graph convolution, our IGC demonstrates three main differences:
\begin{itemize}[leftmargin=*]
    \item \textbf{Incremental neighborhood aggregation}. Instead of aggregating all neighbors, IGC aggregates the new neighbors only, combining with the target node's old representation $\bm{e}_{i,t-1}^{(l+1)}$. Only the parameters associated with new neighbors are trainable (i.e., the $0$-th layer embeddings), which significantly reduces the retraining cost. 
    \item \textbf{Degree synchronizer}. This module revises the normalization weights for old representation and new neighbors to boost the model performance. It requires to save the accumulated degree for each node, which has a negligible cost equal to increasing the embedding size by 1. 
    \item \textbf{Representation aggregator} $\varphi(\cdot)$. This learnable function combines the constant $\bm{e}_{i,t-1}^{(l+1)}$ and the trained new representations, with the ability to handle the discrepancy between them. It can also compensate some estimation error of the old representation and improve model capability, by only introducing a simple CNN with very few parameters. 
\end{itemize}

\subsection{Colliding Effect Distillation}
\label{subsec: CEI}
IGC is a universal operator which is applicable to most GCN models to calculate user and item representations. We apply IGC to a state-of-the-art collaborative filtering GCN model LightGCN~\cite{LightGCN} as \textit{I-LightGCN}, where we stack $L$ IGC layers and set the final representations of all nodes as the average of their representations at all layers. Formally,
\begin{align}\label{eq:embedding}\small
    \bm{r}_{i,t} = \frac{1}{L} \sum_{l=0}^{L}\bm{e}^{(l)}_{i,t} .
\end{align}
We denote the nodes occur in $I_t$ as active nodes and the remaining as inactive nodes. Accordingly, we denote the representations of active and inactive nodes as $\bm{R}_{Ac,t}$ and $\bm{R}_{In,t}$, respectively. Their old representations are organized into $\bm{R}_{Ac,t - 1}$ and $\bm{R}_{In,t - 1}$ with the same criterion.

As we use the new data $I_t$ to train I-LightGCN, IGC mainly refreshes the representation of active nodes, which will thus face the out-of-date issue on the inactive nodes. The reason is that the parameters correspond to the inactive nodes (\eg node embedding) are not involved in the training procedure (\eg back-propagation). In this light, we consider two directions to properly refresh the representation of inactive nodes: 1) directly injecting the new preference signal from active nodes into the representation of inactive nodes; and 2) including the parameters of inactive nodes into the training objective to indirectly push their representations to be updated. Apparently, the key lies in constructing the connection from new data to the representation of inactive node which is cut off due to discarding of data replay in the incremental training setting. 

\textbf{Direct update.} Towards the target, we devise CED as:
\begin{align}\label{equ:CEI inference}
\small
    \bm{\tilde{r}}_{m,t} = \gamma_1 \cdot \bm{r}_{m,t} + \frac{1-\gamma_1}{K} \cdot \sum_{n\in {\text{KNN}(\bm{R}_{Ac,t - 1}, m, K, \delta)}} \bm{r}_{n,t},
\end{align}
where $m$ is an inactive node in stage $t$ and $\bm{\tilde{r}}_{m,t}$ is the final representation that connects $K$ active nodes. $\text{KNN}(\cdot)$ denotes the nearest neighbor fetching operation which calculates the top-$K$ nearest active nodes to node $m$ according to a distance measure $\delta$ (\eg Euclidean distance) between the old representations $\bm{r}_{m,t - 1}$ and $\bm{R}_{Ac,t - 1}$. $\gamma_1 \in [0, 1]$ is a hyper-parameter that controls the influence of active nodes. Intuitively, CED refreshes the representation of an inactive node with the latest status of active nodes that have shown similar properties in previous stages. For instance, we believe that the inactive user $m$ will exhibit similar short-term preference evolution as her/his similar users shown in stage $t$. We will give a rigorous derivation for the formulation of CED in Section~\ref{subsec:casul view}.

\textbf{Indirect update.} While Eq. (\ref{equ:CEI inference}) updates the representations of inactive nodes, their parameters are still not touched since we construct the training data from $I_t$ only. CED thus also operates the active nodes, which is formulated as:
\begin{align}\small\label{equ:CEI loss}
    \bm{\tilde{r}}_{n,t} = \gamma_2 \cdot \bm{r}_{n,t} + \frac{1-\gamma_2}{K} \cdot \sum_{m\in {\text{KNN}(\bm{R}_{In,t - 1}, n, K, \delta)}} \bm{r}_{m,t}.
\end{align}
In this way, the parameters of inactive nodes are attached to the objective function and updated during the retraining. 

Note that CED is also a universal operator applicable to most GCN models. In this work, we apply both IGC and CED to LightGCN, which is named \textit{CI-LightGCN}. Upon the output of CED, we use the inner product of $\tilde{\bm{r}}_{u,t}$ and $\tilde{\bm{r}}_{i,t}$ to generate the prediction result of one $u,i$ pair:
\begin{align}\label{eq:y}
    \hat{y}_{u,i} = <\tilde{\bm{r}}_{u,t} \cdot \tilde{\bm{r}}_{i,t}> .
\end{align}
Following the original LightGCN paper, we learn the model parameters $\bm{\theta}_t$ by optimizing Eq.~(\ref{eq:loss}) over the new data $I_t$ with the mini-batch Adam~\cite{adam} optimizer. In particular, the parameters to be learned include the embedding of nodes and the CNN filters and the $\beta$ in IGC. Note that it is initialized as $\bm{\theta}_{t - 1}$ in the beginning of stage $t$ retraining (random initialization for $t=0$).
The time complexity is close to the standard fine-tuning since only the new interactions $I_t$ are used to construct graph and training examples. 
Below illustrates the retraining procedure of CI-LightGCN.

\vspace{-8pt}
\begin{algorithm}[h]
 \caption{Retraining CI-LightGCN.}
 \label{alg:train}
 \KwIn{Old parameters $\bm{\theta}_{t-1}$, accumulated degree $\{d_{i, 0\sim t-1}\}$, new interactions $I_{t}$} 
 \KwOut{New parameters $\bm{\theta}_{t}$, recommender result $\bm{Y}_{t}$}
 
 $\bm{\theta}_{t} \leftarrow \bm{\theta}_{t-1}$ \algorithmiccomment{Initialization};
 
 Generate incremental graph $G_t$ from $I_t$;
 
 Calculate $\text{KNN}(\bm{R}_{In,t - 1}, n, K, \delta)$ of active nodes;
 
 \While{Stop condition is not reached}{
  Fetch mini-batch data from $I_t$;
  
  Feed forward active node embeddings by Eq.~(\ref{IGC final}) and get active node representations by Eq.~(\ref{eq:embedding});
  
  Generate new active node representations with inactive nodes by Eq.~(\ref{equ:CEI loss}) for indirect update;
  
  Update $\bm{\theta}_t$ by minimizing Eq.~(\ref{eq:loss}) with Eq.~(\ref{eq:y}); 
 }
 
  Calculate representations of all nodes based on optimized $\bm{\theta}_t$ by Eq.~(\ref{IGC final}) and Eq.~(\ref{eq:embedding});

 Calculate $\text{KNN}(\bm{R}_{Ac,t - 1}, m, K, \delta)$ of inactive nodes;

 Generate final inactive node representations by Eq.~(\ref{equ:CEI inference});

\end{algorithm}
\vspace{-8pt}
 In Algorithm~\ref{alg:train}, line $6$ and $10$ refer to \textit{IGC}, 
 \ie the incremental training of the active nodes. 
 Line $3$, $7$, $11$, and $12$ refer to \textit{CED}, where line $7$ and 
 $12$ correspond to \textbf{indirect update} and \textbf{direct update} of inactive nodes, respectively.

\subsection{The Casual View of CED}
\label{subsec:casul view}
We first conceptually introduce colliding effect~\cite{why-book} with a real-life example. Supposing a college gives scholarships to two types of students with unusual musical talents or high GPA. This corresponds to a causal graph: $M \rightarrow S \leftarrow G$ where $M$, $S$, and $G$ denote musical talents, scholarship, and GPA; $S$ is a collider between $M$ and $G$. Ordinarily, musical talent and GPA are independent, but become dependent given that a student won a scholarship. In such a case, knowing that the student lacks musical talent, we can infer that the student is likely to have high GPA. In other words, $M$ and $G$ become dependent by conditioning on the value of their common cause, \ie the collider $S$. In this light, we can leverage such colliding effect to enhance the prediction of $M$ or $G$.

\begin{figure}[t]
	\centering
	\includegraphics[width=0.48\textwidth]{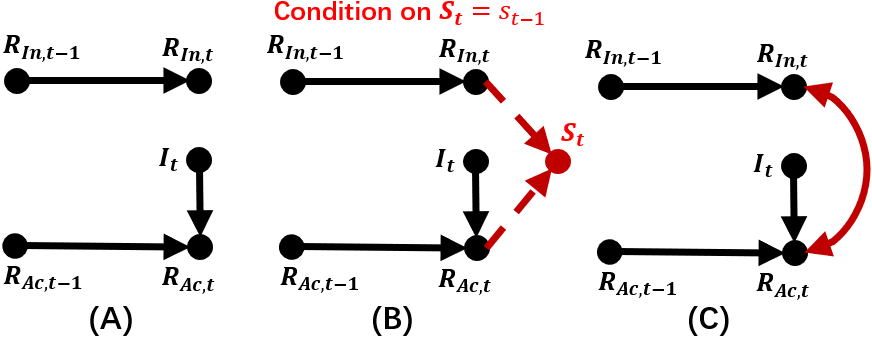}
	\vspace{-3pt}
	\caption{\textbf{(A)} is the causal graph of I-LightGCN to generate node representations in stage $t$. \textbf{(B)} adds an auxiliary variable, the distance $\bm{S}_{t}$ between active and inactive nodes. \textbf{(C)} is the causal graph of CI-LightGCN conditioned on $\bm{S}_{t}=s_{t-1}$.}
	\vspace{-12pt}
    \label{fig:causal graph}
\end{figure}

The design of CED is inspired by~\cite{Distilling}, which first models and leverages the colliding effect in class incremental learning. 
We use three causal graphs~\cite{causal-graph} (see Fig. \ref{fig:causal graph}) to elaborate the causal theory behind CED. We model the representation calculation of I-LightGCN (Fig.~\ref{fig:causal graph}(A)), then introduce the collider to make inactive nodes and new data conditionally $d$-connected  (Fig.~\ref{fig:causal graph}(B)), followed by the causal graph of CI-LightGCN (Fig.~\ref{fig:causal graph}(C)). 
The nodes in a causal graph denote variables, and we scrutinize the meaning of each variable as follow:

\begin{itemize}[leftmargin=*]
    \item $\bm{R}_{In,t}$ and $\bm{R}_{Ac,t}$ denote the node representations of inactive nodes and active nodes, respectively. 
    \item $\bm{R}_{In,t-1}$ and $\bm{R}_{Ac,t-1}$ denote the corresponding old node representations in stage $t-1$. 
    \item $\bm{I}_{t}$ is the new interaction data that is collected in stage $t$.
    \item $\bm{S}_{t}$ denotes the pair-wise distance between nodes in $\bm{R}_{In,t}$ and $\bm{R}_{Ac,t}$.
\end{itemize}
The edges in the causal graph describe the causal relations between variables, where black arrows correspond to the operations in I-LightGCN; red dotted arrows correspond to the calculation of node distance; and double arrows denotes the causal relations as conditioned on a collider. 
In particular,
\begin{itemize}[leftmargin=*]
    \item ($\bm{R}_{Ac,t-1}$, $\bm{I}_{t}$) $\rightarrow$ $\bm{R}_{Ac,t}$: As to the active nodes, IGC aggregates the old representations $\bm{R}_{Ac,t-1}$ and the incremental graph constructed from $\bm{I}_{t}$.
    \item $\bm{R}_{In,t-1} \rightarrow \bm{R}_{In,t}$: As to inactive nodes, IGC only encodes their old representations $\bm{R}_{In,t-1}$. 
    \item ($\bm{R}_{Ac,t}$, $\bm{R}_{In,t}$) $\rightarrow$ $\bm{S}_{t}$: The calculation of $\bm{S}_{t}$ is based on the new representations of active and inactive nodes. Note that $\bm{S}_{t}$ is a collider between variable pairs $\bm{R}_{Ac,t}$ and $\bm{R}_{In,t}$. 
    \item $\bm{R}_{Ac,t}$ $\leftrightarrow$ $\bm{R}_{In,t}$: When conditioned on the collider as $\bm{S}_{t} = s_{t-1}$, the causal path between its parent nodes is built up~\cite{Distilling}.
\end{itemize}

Recall that the key to refreshing the representation of inactive nodes is connecting $\bm{R}_{In,t}$ to $\bm{I}_{t}$. As shown in Fig.~\ref{fig:causal graph}(C), $\bm{I}_{t}$ has colliding effect (CE) on $\bm{R}_{In,t}$ through the path $(\bm{I}_t, \bm{R}_{Ac,t-1}) \rightarrow \bm{R}_{Ac,t} \leftrightarrow \bm{R}_{In,t}$ as conditioned on $\bm{S}_{t}=s_{t-1}$.

$s_{t-1}$ represents the old similarity between active and inactive nodes calculated from $\bm{R}_{Ac,t-1}$ and $\bm{R}_{In,t-1}$. 
Condition on $\bm{S}_{t}=s_{t-1}$ means the updated representations in stage $t$ maintain a similar relative distance between nodes as the previous stage. In this light, we additionally consider the colliding effect $\text{CE}_{\bm{I}_t,\bm{R}_{Ac,t-1}}$, which equals to,
\begin{align}
\label{eq:effect}
&P\left ( \bm{R}_{In,t}| \bm{R}_{In,t-1}, \bm{R}_{Ac,t-1}, \bm{I}_t, \bm{S}_{t} = s_{t-1} \right )\\ \notag
&- P\left ( \bm{R}_{In,t}| \bm{R}_{In,t-1}, \bm{R}_{Ac,t-1}=\bm{0}, \bm{I}_t=\O, \bm{S}_{t} = s_{t-1}\right ),
\end{align}
which denotes the change of $\bm{R}_{In,t}$ as $\bm{R}_{Ac,t-1}$ and $\bm{I}_t$ changes from a reference status ($\bm{R}_{Ac,t-1}=\bm{0}$ and $\bm{I}_t = \O$) to the factual status. We omit the second term since it can be treated as constant. This is because $\bm{R}_{In,t}$ will not be updated if the new data is empty.
By extending the first term according to the total probability formula, we derive $\text{CE}_{\bm{I}_t,\bm{R}_{Ac,t-1}}$ as:
\begin{align}
\label{eq:effect2}
&\sum_{\bm{R}_{Ac,t}}P\left ( \bm{R}_{In,t}| \bm{R}_{In,t-1}, \bm{R}_{Ac,t-1}, \bm{I}_t, \bm{S}_{t}=s_{t-1}, \bm{R}_{Ac,t}\right ) \notag \\ &~~~~~~~ P\left ( \bm{R}_{Ac,t}| \bm{R}_{Ac,t-1}, \bm{I}_t\right )\\
&\ = \sum_{\bm{R}_{Ac,t}} \underline{ P\left ( \bm{R}_{In,t}| \bm{I}_t, \bm{S}_{t}=s_{t-1}, \bm{R}_{Ac,t}\right )} P\left ( \bm{R}_{Ac,t}| \bm{R}_{Ac,t-1}, \bm{I}_t\right )\notag \\
&\ = \sum_{\bm{R}_{Ac,t}} W\left (\bm{I}_t, \bm{R}_{In,t}, s_{t-1}, \bm{R}_{Ac,t}\right ) P\left ( \bm{R}_{Ac,t}| \bm{R}_{Ac,t-1}, \bm{I}_t\right ). \notag
\end{align}
Note that we can omit the variables $\bm{R}_{In,t-1}$, $\bm{R}_{Ac,t-1}$ for briefness (the second step) when conditioned on $\bm{S}_{t}=s_{t-1}$, which is calculated by $\bm{R}_{Ac,t-1}$ and $\bm{R}_{In,t-1}$. By abstracting the underlined term as a weighting function $W\left (\bm{I}_t, \bm{R}_{In,t}, s_{t-1}, \bm{R}_{Ac,t}\right )$, we can understand $\text{CE}_{\bm{I}_t,\bm{R}_{Ac,t-1}}$ as a weighted adjustment of the conditional probability distribution of $\bm{R}_{Ac,t}$. The value of $W$ reflects the important of an active node $\bm{r}_{i,t} \in \bm{R}_{Ac,t}$ to maintain its old similarity ($s_{t-1}$) to an inactive node $\bm{r}_{m,t} \in \bm{R}_{In,t}$. 

Considering that $\bm{R}_{In,t}$ is directly affected by $\bm{R}_{In, t-1}$ (denoted as $\text{DE}_{\bm{R}_{In,t - 1}}$), we estimate their total effect to update the representation of inactive nodes, which is formulated as: $\gamma_1 \cdot \text{DE}_{\bm{R}_{In,t - 1}} + (1 - \gamma_1) \cdot \text{CE}_{\bm{I}_t,\bm{R}_{Ac,t-1}}$). Given an inactive node $m$, we can infer $\text{DE}_{\bm{R}_{In,t - 1}}$ from the output of I-LightGCN (\ie $\bm{r}_{m,t}$). 
As to $\text{CE}_{\bm{I}_t,\bm{R}_{Ac,t-1}}$, we follow~\cite{Distilling} to infer a representation from the weighted distribution according to the $K$-nearest neighbors of $m$ in $s_{t-1}$, which is formulated as:
\begin{equation}
\label{eq:final effect}
    \sum_{n\in {\textit{KNN}(\bm{R_{Ac,t - 1}}, m, K, \delta)}} W \left (\bm{r}_{m,t}, s_{t-1}, \bm{r}_{n,t} \right ) \cdot \bm{r}_{n,t}.
\end{equation}
We omit $\bm{I}_t$ since no co-occurrence of node $m$ and $n$ in $\bm{I}_t$. 
As the simplest average weighting can achieve competitive performance~\cite{Distilling}, we set $W$ as $1/K$. We thus obtain the formulation of CED in Eq.~(\ref{equ:CEI inference}). Similarly, by considering the colliding effect from $\bm{R}_{In, t-1}$ to $\bm{R}_{Ac, t}$, we obtain Eq.~(\ref{equ:CEI loss}).

\section{EXPERIMENTS}
\label{sec:experiment}
In this section, we evaluate \textit{CI-LightGCN} on three real-world datasets to answer the following questions:
\begin{itemize}[leftmargin=*]
    \item \textbf{RQ1}: How is the performance of CI-LightGCN compared with the existing retraining methods?
    \item \textbf{RQ2}: How do the \textit{CED} and \textit{IGC} operators affect the recommendation performance?
    \item \textbf{RQ3}: What factors (\eg hyper-parameters) significantly affect the recommendation performance of \textit{CI-LightGCN}?

\end{itemize}

\vspace{-1.5pt}
\subsection{Experimental Settings}
\label{sec:Experimental Settings}
 \paragraph{Datasets.}
\label{sec:datasets}
We select three widely used datasets for recommendation: Yelp, Gowalla, and Adressa. 1) Yelp is adopted from the 2019 Yelp Challenge, which records the interactions between customers and local business in a period of more than 10 years. 
2) Gowalla\footnote{\url{http://snap.stanford.edu/data/loc-gowalla.html}.} includes the check-in records in one year on go.gowalla.com, where users share their locations by checking in. 3) Adressa is a news articles' clicks records dataset with historical interactions on Adressa in three weeks. For Yelp and Adressa, we adopt the version\footnote{\url{https://github.com/zyang1580/SML/}.} pre-processed by~\cite{SML}, where the interactions are chronologically split into stages. For Gowalla, we follow the same pre-processing procedure and split the interactions into 40 stages. We summarize detailed statistics of the datasets in TABLE~\ref{table:dataset}. And the densities of Yelp, Gowalla, Adressa are 0.042\%, 0.084\%, and 0.037\%, respectively. We evaluate the effectiveness of top-$K$ recommendation\footnote{The code of our method CI-LightGCN is available at  \url{https://github.com/Dingseewhole/CI_LightGCN_master/}} by reporting the average recall@$5, 20$ (R@5 and R@20) and ndcg@$5, 20$ (N@5 and N@20) over interactions in the testing stages.

\begin{table}[]
\centering
    \caption{Statistics of the three used datasets.}
    \vspace{-0.3cm}
    \label{table:dataset}
    \setlength{\tabcolsep}{1.5mm}{
    \resizebox{0.5\textwidth}{!}{%
    \begin{tabular}{@{}lllllll@{}}
    \toprule
    Dataset  & \#Users & \#Items & \#Interactions & \#Train & \#Validation   & \#Test           \\ \midrule
    Yelp & 122,816  & 59,082   & 3,014,421 & {[}0-30) stage & {[}30-33) stage & {[}33-39{]} stage \\
    Gowalla  & 29,858   & 40,981   & 1,027,370 & {[}0-30) stage & {[}30-33) stage & {[}33-39{]} stage \\
    Adressa  & 478,612  & 20,875   & 3,664,225 & {[}0-48) stage & {[}48-53) stage & {[}53-62{]} stage \\ 
    \bottomrule
    \end{tabular}
    }
    }
    \vspace{-8pt}
\end{table}

\begin{table*}[htbp]
\centering
\caption{Recommendation performance on Yelp and Gowalla. The best performance and best baseline in each column are highlighted with bold font and underline, respectively. RI denotes \textbf{CI-LightGCN}'s relative performance gain \wrt R@5.}
\label{table:performance}
\vspace{-0.2cm}
\setlength{\tabcolsep}{1.5mm}{
\resizebox{0.85\textwidth}{!}{%
\begin{tabular}{l|cccc|c|cccc|c}
\hline
{Methods}               & \multicolumn{5}{c|}{{Yelp}}                                                                                                                           & \multicolumn{5}{c}{{Gowalla}} \\                                  
 & \multicolumn{1}{c}{R@5} & \multicolumn{1}{c}{R@20} & \multicolumn{1}{c}{N@5} & \multicolumn{1}{c}{N@20} & RI & \multicolumn{1}{c}{R@5} & \multicolumn{1}{c}{R@20} & \multicolumn{1}{c}{N@5} & \multicolumn{1}{c}{N@20} & RI \\ \hline \hline
\textbf{GRU4Rec} & 0.1706 & 0.4158 & 0.1080 & 0.1771 & 92.5\% & 0.1016  & 0.2978  & 0.0623  & 0.1169  & 215.8\% \\ 
\textbf{Caser} & 0.2195 & 0.4565 & 0.1440 & 0.2117 & 49.6\% & 0.1143  & 0.3622  & 0.1111 & 0.3585  & 180.8\% \\ 
\textbf{SPMF}  & 0.1725  & 0.3635  & 0.1136 & 0.1677 & 90.4\% & 0.1595 & 0.3759 & 0.0993 & 0.1606  & 101.2\%    \\ 
\textbf{SML-MF}  & 0.2251  & 0.4748 & 0.1485 & 0.2194 & 45.9\%  & 0.1761  & 0.3428 & 0.1233 & 0.1701 & 82.2\%    \\ \hline
\textbf{Fine-tune LightGCN} & 0.2338 & 0.4320 & 0.1619 & 0.2185 & 40.5\% & 0.2142 & 0.3860 & 0.1505 & 0.1993 & 49.8\% \\
\textbf{Full-retrain LightGCN} & {\underline{0.2923}} & {\underline{0.5247}} & {\underline{0.2074}} & {\underline{0.2727}} & 12.4\% & {\underline{0.3103}} & {\underline{0.5276} } & {\underline{0.2212}} & {\underline{0.2819}} & 3.4\% \\ 
\textbf{SML-LightGCN-O} & 0.1895 & 0.4197 & 0.1246 & 0.1897 & 73.3\% & 0.2152 & 0.4275 & 0.1505 & 0.2103 & 49.1\% \\ 
\textbf{SML-LightGCN-E} & 0.1771 & 0.3971  & 0.1159 & 0.1782 & 85.4\% & 0.2153 & 0.4515 & 0.1476 & 0.2144 & 49.0\% \\ 
\textbf{LightGCN+EWC} & 0.2365 & 0.4441  & 0.1736 & 0.2463 & 38.9\% & 0.2117 & 0.3942 & 0.1498 & 0.1987 & 51.6\% \\\hline
\textbf{CI-LightGCN}  & \textbf{0.3284}  & \textbf{0.5695} & \textbf{0.2294} & \textbf{0.2956} &- & \textbf{0.3209} & \textbf{0.5421}  & \textbf{0.2272} & \textbf{0.2908}  &-    \\ \hline
\end{tabular}%
}
}
\vspace{-0.4cm}
\end{table*}

\paragraph{Compared methods.}
\label{sec:baseline}
We consider five LightGCN models with different retraining methods:

\begin{itemize}[leftmargin=*]
    \item \textbf{Full-retrain LightGCN}. This method retrains LightGCN with all interactions and full graph. We search the L2-norm coefficient in $\left [0, 0.001\right]$ at a multiplicative ratio of 10x; training epochs in $\left [400, \text{1,000} \right ]$ with the step of $100$.
    \item \textbf{Fine-tune LightGCN}. This method updates LightGCN with new interactions and the incremental graph\footnote{We omit the version of fine-tuning with old graph, which is less effective than full retraining and less efficient than fine-tuning w/o old graph.}. We search the L2-norm coefficient in $\left [0, 0.001 \right ]$ at a multiplicative ratio of 10x; training epochs in $\left \{200, 300, 400, 500 \right \}$.
    \item \textbf{SML+LightGCN-O}. This method retrains LightGCN with SML~\cite{SML}, which firstly trains a LightGCN over the incremental graph, and then combines the old and new representations with a Transfer. We search the L2-norm coefficient in $\left [0, 0.01\right ]$ at a multiplicative ratio of 10x; training epochs of LightGCN and Transfer in $\left [50, 600\right ]$ with the step of $50$; learning rate in $\left [ 1e\text{-}5, 1e\text{-}2 \right ]$ at 10x multiplicative ratio.
    \item \textbf{SML+LightGCN-E}. It also retrains LightGCN with SML, which combines the parameters (\ie the $0$-th layer's node embedding) of the new model and old model. We search the same hype-parameters as \textbf{SML+LightGCN-O}. 
    \item \textbf{LightGCN+EWC}. This method retrains LightGCN with Elastic Weight Consolidation (EWC)~\cite{EWC} over the incremental graph and new interactions, which is a regularization-based method in continual learning. We search the EWC ratio in $\left [0.01, 1\right ]$ at a multiplicative ratio of 10x.
\end{itemize}
We also include some sequential recommendation baselines.
\begin{itemize}[leftmargin=*]
    \item \textbf{GRU4Rec}~\cite{GRU4rank}. It pioneered the usage of recurrent neural network (RNN) to serve the session-based recommender system. By building a RNN for each user's interaction sequence, it can capture the interest drift of users. We use the whole history to construct the user's interaction sequence, and retrain the model with full-retrain strategy. We search the hidden layer size of GRU in $\left \{64, 128, 256\right \}$.
    \item \textbf{Caser}~\cite{caser}. This method uses CNN to capture the interest evolution of users and the collaboration signal between items of the most $L$ recent interactions. We tune $L$ in the range of $[1, 5]$ with step $1$, and other hyper-parameters follow the optimal setting as reported in the paper.
    \item \textbf{SPMF}~\cite{spmf}. It applies a sample-based retraining method on Matrix Factorization (MF), which samples some historical interactions to be added into new data to update the old model. We search the best reservoir size in $\left \{7,000,15,000,30,000,70,000,150,000 \right \}$.
    \item \textbf{SML-MF}~\cite{SML}. This method applies SML on MF where the Transfer combines the old embedding and new embedding. We quote the result from SML~\cite{SML}. 
\end{itemize}

\vspace{-2pt}
\subsection{Performance Comparison (RQ1)}
\label{sec:Performance Comparison}
TABLE~\ref{table:performance} reports the performance comparison results. From the table, we have the following observations:
\begin{itemize}[leftmargin=*]
    \item CI-LightGCN outperforms Full-retrain LightGCN and Fine-tune LightGCN in all cases. It validates the rationality of combining the old representation and incremental graph to approach the full graph convolution. The performance gain of CI-LightGCN is attributed to the IGC and CED, which properly fuse the long-term and short-term preference for all users and items. 
    \item CI-LightGCN also outperforms SML-LightGCN-O and SML-LightGCN-E, which combine the old and new representations with the complex Transfer model. We postulate the reasons are twofold: 1) the complex Transfer model hurts the performance of LightGCN due to its feature transformation and nonlinear activation~\cite{LightGCN}; and 2) the SML-based methods use degree within incremental graph only for normalization weighting, which ignore the accumulated degree. This result indicates the importance of normalization weighting in GCN model, and the importance of synchronizing the weights for GCN retraining. 
    \item LightGCN+EWC beats Fine-tune LightGCN since EWC alleviates the forgetting issue of old preference signal. Nevertheless, CI-LightGCN further achieves performance gain over LightGCN+EWC. The reason for this is due to the IGC and CED operators. IGC transfers old representations into new representations, whereas EWC just treats it as a regularizer, and CED updates the representation of inactive nodes but EWC is unable to do that.
    \item On all datasets, LightGCN-based methods largely outperform the non-LightGCN ones in most cases, including SML-MF equipped with advanced retraining strategy. It is consistent with the result in~\cite{NGCF}, validating the effectiveness of GCN in recommendation.
\end{itemize}

\begin{figure}[t]
    \centering
 	\vspace{-4pt}
	\includegraphics[width=0.55\textwidth]{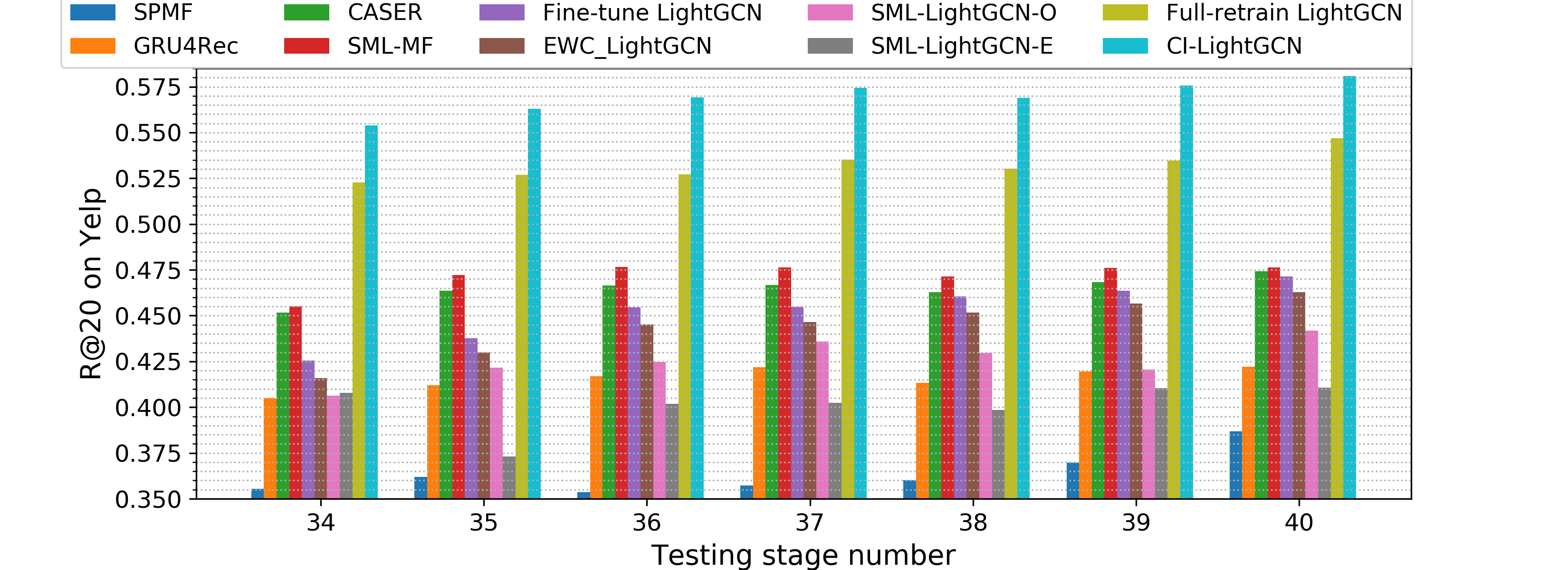}
 	\vspace{-10pt}
	\caption{Stage-wise performance on Yelp.}
	\label{fig:timing result}
	\vspace{-4pt}
\end{figure}

\paragraph{Stage-wise Performance}
Fig.~\ref{fig:timing result} shows the detailed
recommendation performance \wrt by R@10 at each testing stage of Yelp. To save space, we omit the results of other
metrics and the results on Yelp, which show the same trend. From the figure, we can see that CI-LightGCN stably outperforms the baselines across the stages.

\begin{figure}[htbp]
    \centering
	\vspace{-11pt}
	\includegraphics[width=0.42\textwidth]{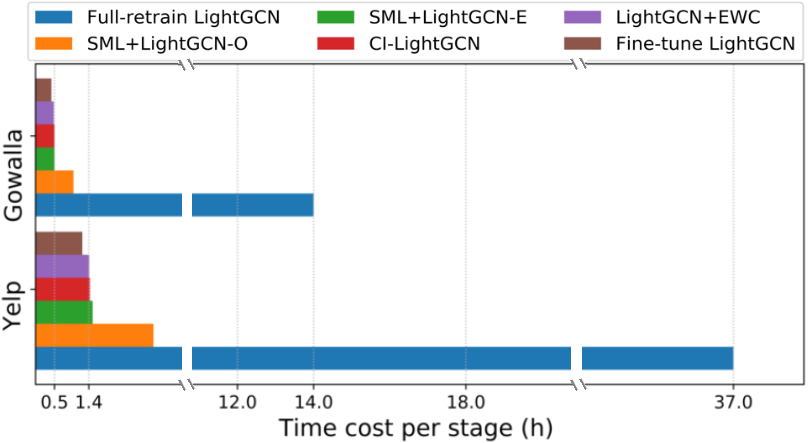}
 	\vspace{-7pt}
	\caption{Training time of different LightGCN-based methods.}
	\label{fig:time cost}
	\vspace{-15pt}
\end{figure}

\paragraph{Speed-up}
Recall that our target is the efficient retraining of GCN model, we compare the time cost of LightGCN-based methods. Fig.~\ref{fig:time cost} shows their training time on the same server with one RTX-3090 GPU. From the figure, we can see that: 1) CI-LightGCN speeds up the retraining with more than 30 times compared with full-retraining; and 2) the running time of CI-LightGCN is slightly longer than fine-tuning.
The results justify that IGC and CED enable the fast retraining of GCN model, which is highly valuable in practice.

\subsection{Ablation Study (RQ2)}\label{sec:ab study}

\paragraph{Study on CED} To reveal the effect of CED and IGC, we further evaluate two variants of CI-LightGCN: 1) I-LightGCN, which only applies IGC on LightGCN, \ie removing CED; and 2) CI-LightGCN(T), which only uses CED in training, \ie removing Step 11 and 12 of Algorithm~\ref{alg:train}. TABLE~\ref{table:study_ced} shows their recommendation performance on the two datasets. From the table, we can see that the recommendation performance of the three methods exhibit a clear increase trend, which validates the rationality of updating inactive nodes during incremental training and the effectiveness of CED. Moreover, I-LightGCN still outperforms the baselines in TABLE~\ref{table:performance}, which justifies the effectiveness of IGC. Furthermore, we evaluate the detailed performance on inactive users as the target of CED is to refresh their representations. Across TABLE~\ref{table:inactive node} and~\ref{table:study_ced}, we observe that CI-LightGCN(T) and CI-LightGCN achieve larger improvement over I-LightGCN on inactive users. It validates that CED can boost the recommendation accuracy of inactive users.

\begin{table}[t]
\centering
\caption{Performance of CI-LightGCN and its variants.}
\vspace{-0.2cm}
\label{table:study_ced}
\begin{tabular}{l|l|cccc}
\hline
Datasets  & Methods  & R@5   & R@20  & N@5  & N@20 \\ \hline
     & I-LightGCN           & 0.3222  & 0.5566 & 0.2261 & 0.2933 \\
Yelp & CI-LightGCN(T)   & 0.3249 & 0.5663 & 0.2262 & 0.2954  \\
     & CI-LightGCN & \textbf{0.3284}  & \textbf{0.5695} & \textbf{0.2294} & \textbf{0.2956} \\ \hline
     & I-LightGCN           & 0.3173 & 0.5369  & 0.2248 & 0.2878 \\
Gowalla & CI-LightGCN(T)   & 0.3179 & 0.5370  & 0.2260 & 0.2890 \\
        & CI-LightGCN & \textbf{0.3209} & \textbf{0.5421}  & \textbf{0.2272} & \textbf{0.2908} \\ \hline
\end{tabular}
\end{table}

\begin{table}[t]
\centering
\caption{Performance of CI-LightGCN and its variants on inactive users.}
\vspace{-0.2cm}
\label{table:inactive node}
\begin{tabular}{l|l|cccc}
\hline
Datasets  & Methods  & R@5   & R@20  & N@5  & N@20 \\ \hline
     & I-LightGCN           & 0.2712 & 0.4724 & 0.1925 & 0.2507 \\
Yelp & CI-LightGCN(T)   & 0.2783 & 0.4864 & 0.1945 & 0.2542  \\
     & CI-LightGCN & \textbf{0.2852} & \textbf{0.4959} & \textbf{0.1994} & \textbf{0.2598} \\ \hline
     & I-LightGCN           & 0.2807 & 0.4510 & 0.2026 & 0.2515 \\
Gowalla & CI-LightGCN(T)   & 0.2839 & 0.4549 & 0.2063 & 0.2543 \\
        & CI-LightGCN & \textbf{0.2870} & \textbf{0.4599} & \textbf{0.2091} & \textbf{0.2587} \\ \hline
\end{tabular}
\vspace{-10pt}
\end{table}

\begin{figure}[t]
    \centering
	\includegraphics[width=0.45\textwidth]{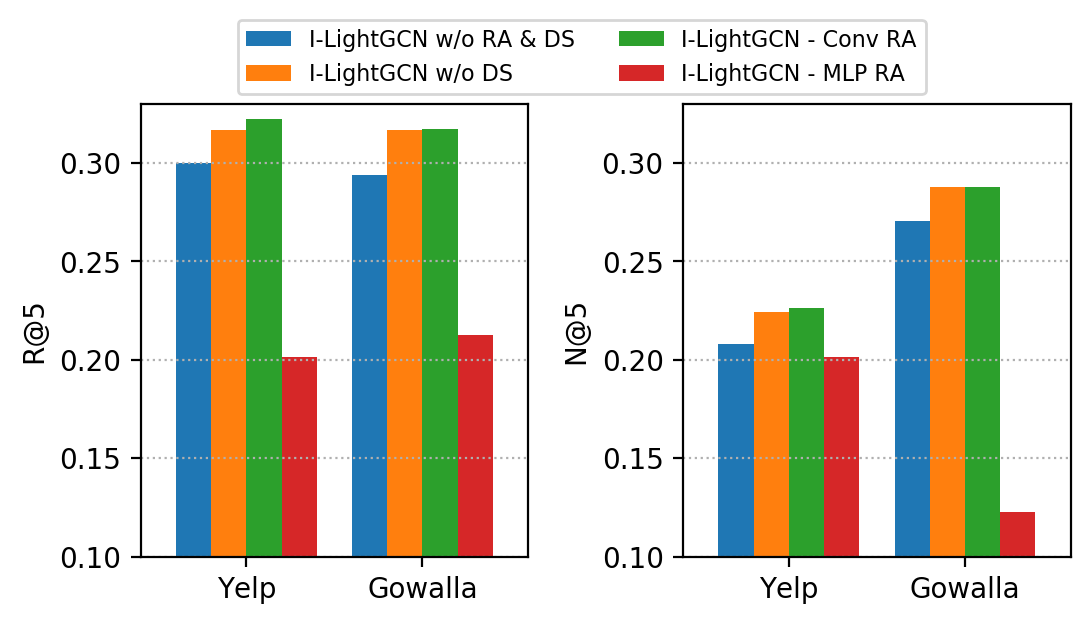}
 	\vspace{-10pt}
	\caption{Performance of I-LightGCN and its variants.}
	\label{fig:depth study of IGC-LightGCN}
	\vspace{-15pt}
\end{figure}
\paragraph{Study on IGC} We study how the components of IGC influence its effectiveness by comparing two variants of I-LightGCN without degree synchronizer (I-LightGCN w/o DS), and without representation aggregator as well (I-LightGCN w/o RA \& DS). Meanwhile, we compare the I-LightGCN - MLP RA which enables the parameter sharing across dimensions in the representation aggregator. Fig.~\ref{fig:depth study of IGC-LightGCN} shows the performance comparison between vanilla I-LightGCN and its variants. From the figure, we can see that: 1) The performance of vanilla I-LightGCN (\ie I-LightGCN - Conv RA), I-LightGCN w/o DS, and I-LightGCN w/o RA \& DS show a clear decrease trend in most cases, which justifies the effectiveness and the necessity of the two modules in IGC; 2) I-LightGCN - MLP RA performs much worse than the one with CNN implementation across the two datasets, which demonstrates the rationality of restricting the complexity of the representation aggregator.

\subsection{In-depth Analysis (RQ3)}

\begin{figure}[t]
\centering
\subfigure{
\begin{minipage}[t]{0.49\linewidth}
\centering
\includegraphics[width=0.98\textwidth]{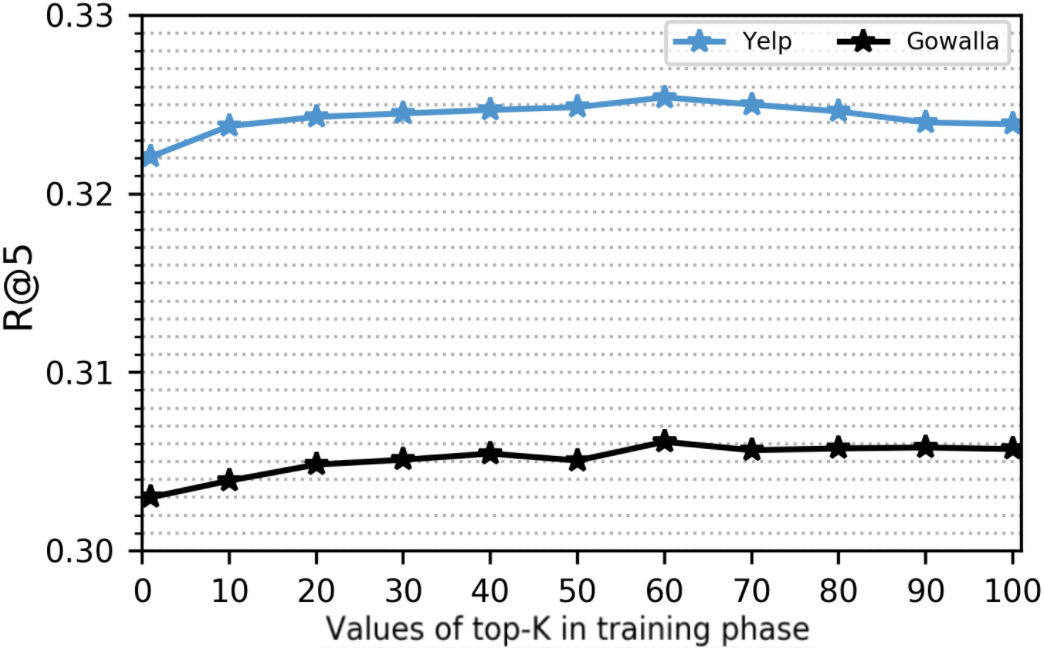}
\end{minipage}%
}%
\subfigure{
\begin{minipage}[t]{0.50\linewidth}
\centering
\includegraphics[width=1\textwidth]{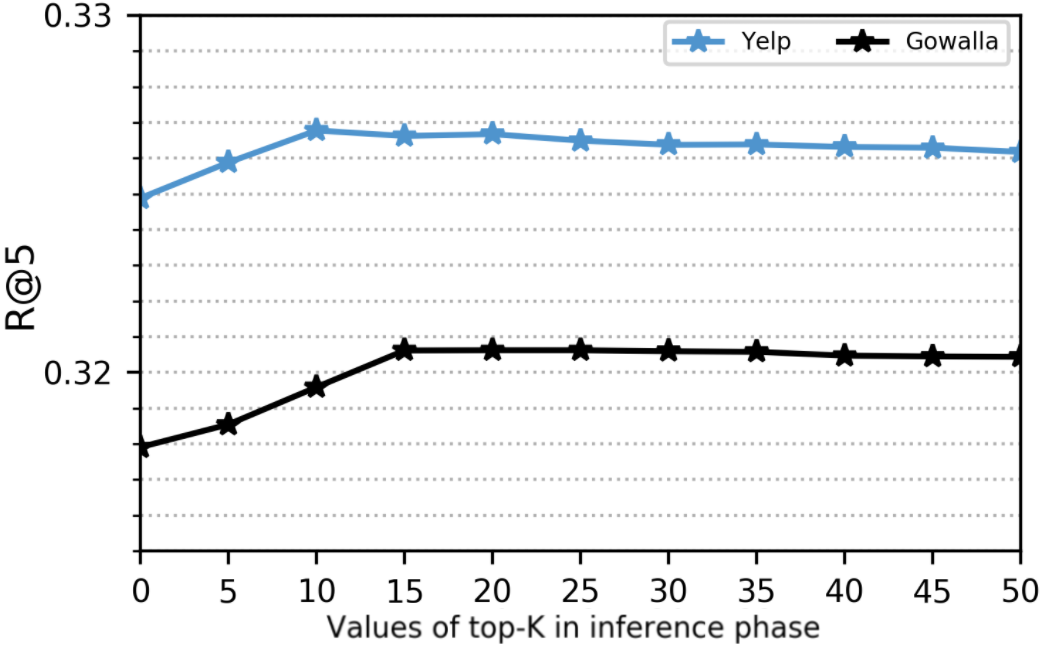}
\end{minipage}%
}%
\centering
\caption{Performance of CI-LightGCN as changing the top-$K$ in training phase (\textbf{left}) and inference phase (\textbf{right}).}
\label{fig:k train-inference}
\end{figure}

\begin{figure}[t]
\centering
\subfigure{
\begin{minipage}[t]{0.49\linewidth}
\centering
\includegraphics[width=0.98\textwidth]{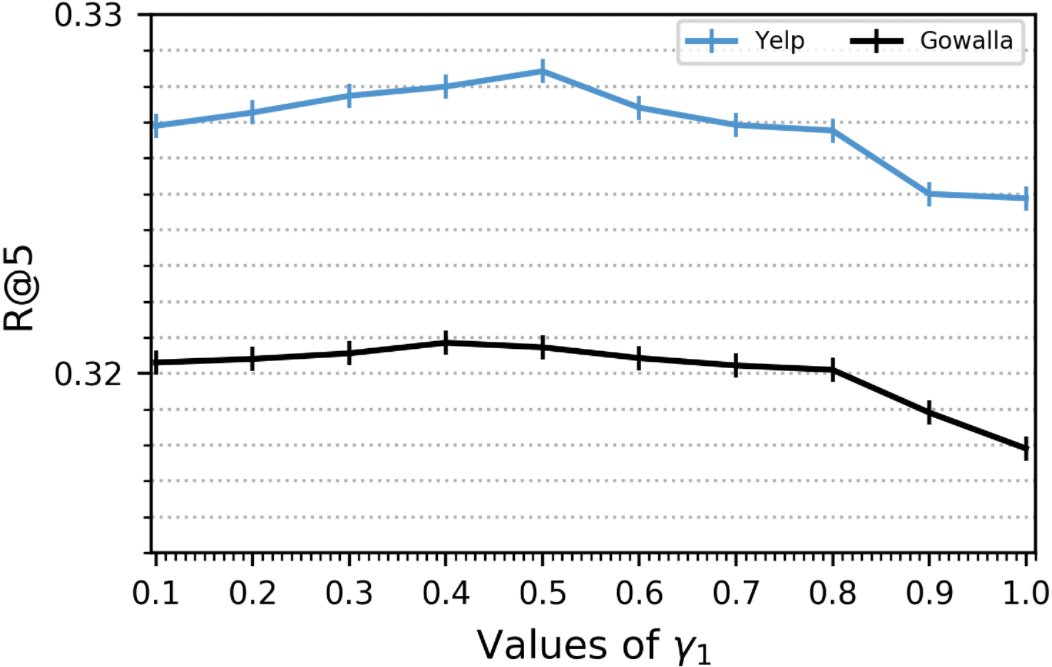}
\end{minipage}%
}%
\subfigure{
\begin{minipage}[t]{0.49\linewidth}
\centering
\includegraphics[width=0.98\textwidth]{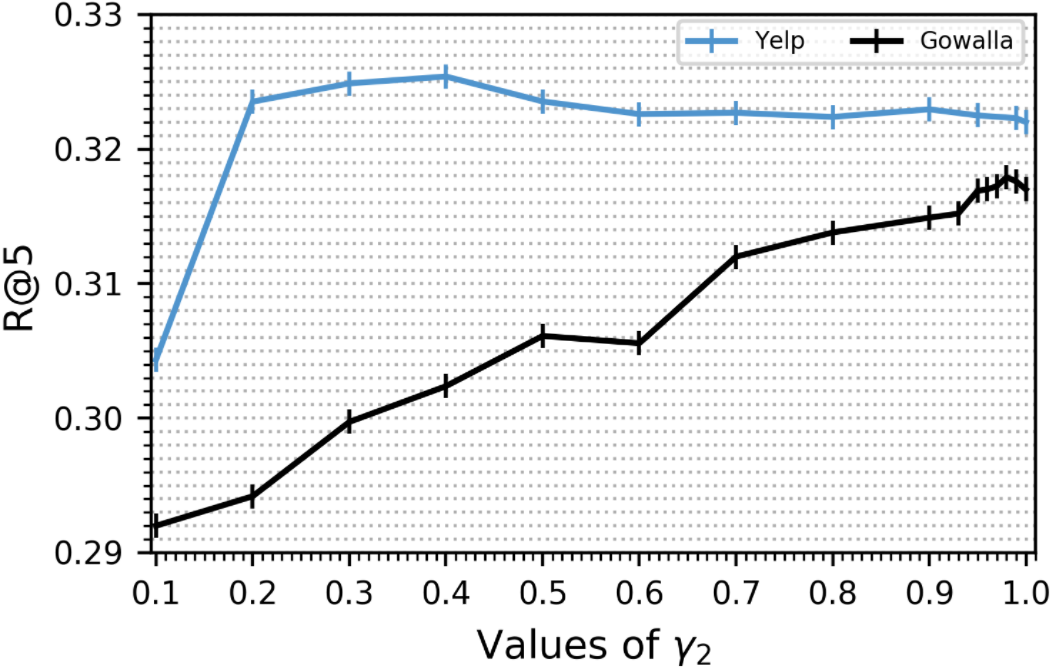}
\end{minipage}%
}%
\centering
\vspace{-10pt}
\caption{Performance of CI-LightGCN as changing the value of $\gamma_1$ (\textbf{left}) and $\gamma_2$ (\textbf{right}).}
\vspace{-10pt}
\label{fig:alpha beta}
\end{figure}

\paragraph{Study on CED}
We investigate how the hyper-parameters influence the effectiveness of CED. We select four hyper-parameters: 
the value of $K$ and $\gamma$ (\ie $\gamma_1$ and $\gamma_2$)
in Eq. (\ref{equ:CEI loss}) and (Eq. \ref{equ:CEI inference}).
Fig.~\ref{fig:k train-inference} and~\ref{fig:alpha beta} show the performance of CI-LightGCN as changing the value of $K$ and $\gamma$. 
From Fig.~\ref{fig:k train-inference}, we can observe a similar trend of performance (increasing then stable) across the four curves, \ie CED is insensitive to $K$ and we can simply set $K$ as a relative small value (\eg 15). 
Fig.~\ref{fig:alpha beta} shows that CI-LightGCN performs worst as $\gamma_1=1.0$, which reveals the benefit of colliding effect distillation on inactive nodes. Moreover, the performance largely decreases as setting $\gamma_2$ with small values. This is reasonable since smaller $\gamma_2$ leads to less update of the active nodes during training.

\begin{figure}[t]
    \centering
	\includegraphics[width=0.48\textwidth]{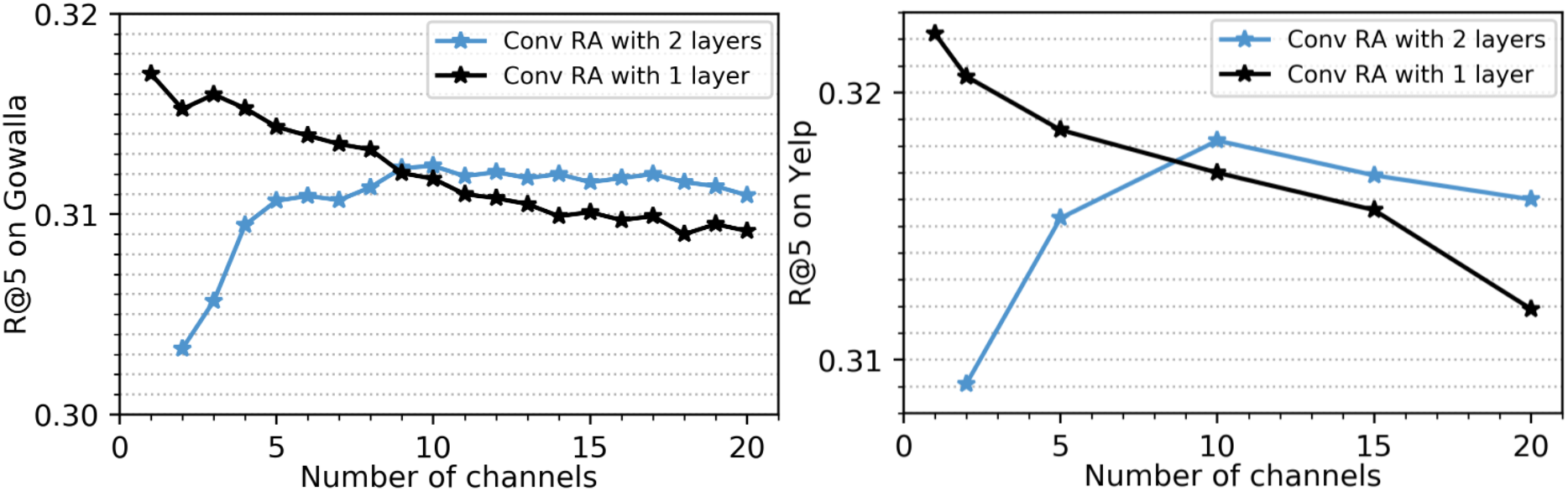}
 	\vspace{-5pt}
	\caption{Performance of I-LightGCN as changing the RA.}
	\label{fig:RA-complexity}
	\vspace{-18pt}
\end{figure}

\paragraph{Study on IGC}
For IGC, we select two hyper-parameters in the representation aggregator: the number of CNN layers and the number of CNN filters, to study the sensitivity of IGC. Fig.~\ref{fig:RA-complexity} shows the performances of I-LightGCN with 1-layer CNN and 2-layer CNN on Gowalla and Yelp as changing the number of filters from 1 to 20. We can see that: 1) the 1-layer CNN performs better than the 2-layer CNN in most cases; and 2) adding more filters will hurt the performance of 1-layer CNN. This result suggests to set the aggregator with only one filter, which means that a weighted combination is sufficient for combining the old and new representations in practical usage.

\paragraph{Study on Time-sensitivity}
As shown in~\cite{SML}, retraining methods can perform distinctly on datasets with different time-sensitivity. We thus further test the LightGCN-based methods on Adressa, which is highly time-sensitive. TABLE~\ref{table:Adressa} shows the performance of these methods. Due to the time-sensitivity, Full-retrain LightGCN and LightGCN + EWC, which highlight the long-term preference signal, perform much worse than Fine-tune LightGCN. Nevertheless, I-LightGCN achieves performance comparable to Fine-tune LightGCN owing to the flexibility brought by the representation aggregator and degree synchronizer in IGC, which enable the model to also approach Fine-tune LightGCN by down weighting the old representation and degree. This result demonstrate the robustness of IGC for different datasets.

\begin{table}[th]
\centering
\caption{Recommendation performance of LightGCN-based methods on Adressa. The best and runner-up in each column are highlighted with bold font and underline, respectively.}
\vspace{-5pt}
\label{table:Adressa}
\begin{tabular}{l|cccc}
\hline
Methods               & R@5            & R@20           & N@5            & N@20           \\ \hline
I-LightGCN          & \textbf{0.254} & \textbf{0.415} & \textbf{0.181} & \textbf{0.227} \\
Full-retrain LightGCN & 0.028          & 0.094          & 0.018          & 0.227          \\
Fine-tune LightGCN    & {\underline{0.253}}    & {\underline{0.412}}    & {\underline{0.181}}    & {\underline{0.227}}    \\
LightGCN + EWC        & 0.252          & 0.393          & 0.177          & 0.218          \\
SML-LightGCN-O        & 0.243          & 0.411          & 0.170          & 0.216          \\
SML-LightGCN-E        & 0.243          & 0.412          & 0.168          & 0.217          \\ \hline
\end{tabular}
\vspace{-20pt}
\end{table}
\section{RELATED WORK}
\label{relacted work}
\textbf{GCN-based Recommendation.} 
In recent years, GCN has become the cutting-edge technique for recommendation~\cite{DBLP:conf/icml/YuQ20,DBLP:conf/kdd/ChenW20,DBLP:conf/kdd/JinQFD00ZS20,DBLP:conf/kdd/XuZZLSG20,yang2020learning,wei2020fast,ipngcn2021}. A surge of attention has been dedicated on designing GCN models to learn comprehensive user and item representations from the interaction graph for collaborative filtering~\cite{NGCF,LightGCN,DBLP:conf/icml/YuQ20}. 
Beyond the user-item interactions, a line of research explores GCN models to encode more types of relations such as item relation, social network~\cite{DBLP:conf/pakdd/WangLWF20,DBLP:conf/wsdm/Song0WCZT19,DBLP:conf/www/Fan0LHZTY19}, and knowledge graph~\cite{DBLP:conf/kdd/Wang00LC19,DBLP:conf/cikm/SunCZWZZWZ20}. 
However, the existing researches focus on model efficacy, largely ignore the efficiency of model training, which is very important in practical usage. In this work, we focus on efficient retraining of GCN model, which is in an orthogonal direction to the existing research.

\textbf{Dynamic GCN.}
Graph is a constantly changing data structure in the real world, such as social network, academic network, point cloud. In response to this issue, plenty of previous work explore Dynamic GCN to capture the temporal feature of graph data~\cite{ctdne,DBLP:conf/ijcai/QiuCA19,DBLP:conf/cikm/WangSWW20,evolvegcn,LiuTKDE2021,Yuanijcai20}, which are mainly in two categories according to viewing the graph evolution in stages or a random process. For instance, DySAT~\cite{DBLP:conf/wsdm/SankarWGZY20} arranges the snapshots of the graph into a sequence with the temporal order, and uses two decoupled modules to encode the structure feature and the temporal feature within the sequence. DyRep~\cite{Dyrep} models the occurrence of an edge as a point process and using a neural network to parameterize the intensity function.

There are also attempts to use Dynamic GCN for recommendation~\cite{DBLP:conf/wsdm/Song0WCZT19,DBLP:conf/kdd/KumarZL19}, so as to capture the temporal pattern of user-item interactions. For instance, JODIE~\cite{DBLP:conf/kdd/KumarZL19} couples an RNN with GCN to learn the dynamic representation of users/items from a sequence of interaction graphs. GraphSAIL~\cite{GraphSAIL} can retrain the graph-based recommender model with global and local structure distillation.
Despite the success of Dynamic GCN in capturing the short-term and long-term preference, it requires long interaction history for training, which is memory and time costly by nature. In this work, we focus on the efficient retraining of GCN model, where only the incremental graph is accessible in our setting. The Dynamic GCN models are thus not applicable here, which are omitted in the experiments for comparison.
Although GraphSAIL~\cite{GraphSAIL} can incremental train the old model, it performs worse than intuitive full-retraining method (named \textit{named Fu\_batch} in~\cite{GraphSAIL} Tab. 3). Since our target is to deign a retraining method that efficient as fine-tuning but without sacrificing the performance, and our proposed IGC and CED always performance better than full-retraining, we omitted this method in the experiments.

\textbf{Causal Recommendation.}
A surge of attention has been dedicated to incorporating causal inference techniques into various machine learning applications~\cite{visual-rcnn,long-tail,10.1145/3404835.3462823,yang2021causal,feng2021empowering,chen2020counterfactual,niu2021counterfactual}. These methods are also successfully adapted to recommendation systems for resolving bias issues and enhancing model reliability. 
For instance, DecRS~\cite{DBLP:journals/corr/abs-2105-10648}, MACR~\cite{wei2020model}, and KDCRec~\cite{liu2020general} analyze the biases of recommendation from causal view and use counterfactual learning techniques to debias. 
Similarly, intervention techniques are also leveraged to debias~\cite{PDA,DBLP:journals/corr/abs-1808-06581, DBLP:conf/recsys/WangLCB20}. Existing works focus on leveraging causal inference techniques for debiasing.
Our work is in an orthogonal direction as we focus on model retraining and colliding effect distillation, which is a new causal inference technique.
\section{CONCLUSION}
\label{sec:con}
In this work, we highlighted the importance of GCN model retraining for recommendation. To achieve effective and efficient retraining, our analysis enlightens that the key lies in detaching the old graph from neighborhood aggregation, meanwhile, reserving the long-term preference signal and refreshing the inactive nodes. Towards this end, we proposed a Causal Incremental Graph Convolution method, which consists of two new operators named Incremental Graph Convolution (IGC) and Colliding Effect Distillation (CED). We instantiated IGC and CED based on LightGCN and conducted extensive experiments on three real-world datasets. The results show that CI-LightGCN outperforms full retraining with a speed-up of more than 30 times, validating the effectiveness and the rationality of our proposal. 

This work opens up a new research direction about GCN model retraining, and highlights a promising perspective about causal inference. In the future, we are interested in applying different model parameter updating techniques in IGC such as meta-learning~\cite{SML}. In addition, we will test IGC and CED in more graph learning applications, such as user profiling and targeted advertising.

\bibliographystyle{IEEEtran}
\bibliography{reference}

\ifCLASSOPTIONcaptionsoff
 \newpage
\fi
\end{document}